\RequirePackage{fix-cm}
\documentclass[smallextended]{svjour3}       
\smartqed  
\usepackage{graphicx}
\usepackage[separate-uncertainty=true,output-open-uncertainty=,output-close-uncertainty=]{siunitx}
\usepackage{url}
\usepackage[colorlinks=true]{hyperref}

\usepackage[utf8]{inputenc}

\usepackage[usenames,dvipsnames,svgnames,table]{xcolor}
\usepackage{array}

\usepackage[caption=false,font=footnotesize]{subfig}

\usepackage{tikz}
\usetikzlibrary{positioning}

\usepackage{bm}

\usepackage{algorithmicx}
\usepackage[noend]{algpseudocode}

\usepackage[%
square,        
comma,         
numbers,       
sort&compress, 
]{natbib}

\newcommand{\etal}{\emph{et.al.}}

\newcommand{\figref}[1]{Fig.~\ref{#1}}
\newcommand{\secref}[1]{Section~\ref{#1}}
\newcommand{\tabref}[1]{Table~\ref{#1}}
\newcommand{\eqnref}[1]{Equation~\ref{#1}}

\newcommand{\orientationangle}{forward angle}

\newcommand{\formulapose}{\boldsymbol{\theta}}
\newcommand{\formulaposition}{\boldsymbol{x}}

\newcommand{\formulainstrumentposition}{\vec{x}_{instr}}
\newcommand{\formulainstrumentorientation}{\vec{n}_{instr}}
\newcommand{\formulaorientangle}{\alpha}

\newcommand{\formulatilt}{\tau}

\newcommand{\formuladepth}{d}

\newcommand{\changed}[1]{#1}

\newcommand{\gauss}[2][0]{\mathcal{N}(#1,#2)}
\newcommand{\uniform}[2]{\mathcal{U}(#1,#2)}

\newcommand{\by}[2]{$#1\mathsf{x}#2$}

\journalname{IJCARS}

\begin{document}

\def\ourtitle{i3PosNet: Instrument Pose Estimation from X-Ray in temporal bone surgery}
\title{\ourtitle
}



\author{David~Kügler \and
        Jannik~Sehring \and
        Andrei~Stefanov \and
        Igor~Stenin \and
        Julia~Kristin \and
        Thomas~Klenzner \and
        Jörg~Schipper \and
        Anirban~Mukhopadhyay%
}


\institute{D. Kügler \at
	                German Center for Neuro-degenerative Diseases, Bonn, Germany\\
	                \emph{work done at} Department of Computer Science, TU Darmstadt, Darmstadt, Germany\\
	\email{david.kuegler@dzne.de}           
	                \and
	                J. Sehring \and A. Mukhopadhyay \at
	Department of Computer Science, Technischer Universität Darmstadt, Darmstadt, Germany\\
	           \and
	           I. Stenin \and J. Kristin \and T. Klenzner \and J. Schipper \at
	ENT Clinic, University Düsseldorf, Germany
}

\date{Received: date / Accepted: date}

\maketitle

\begin{abstract}
~\par
\textbf{Purpose}:
Accurate estimation of the position and orientation (pose) of surgical instruments is crucial for delicate minimally invasive temporal bone surgery.
Current techniques lack in accuracy and/or line-of-sight constraints (conventional tracking systems) or expose the patient to prohibitive ionizing radiation (intra-operative CT).
A possible solution is to capture the instrument with a c-arm at irregular intervals and recover the pose from the image.

\textbf{Methods}:
i3PosNet infers the position and orientation of instruments from images using a pose estimation network.
Said framework considers localized patches and outputs pseudo-landmarks.
The pose is reconstructed from pseudo-landmarks by geometric considerations.

\textbf{Results}:
We show i3PosNet reaches errors less than \SI{0.05}{\milli\meter}.
It outperforms conventional image registration-based approaches reducing average and maximum errors by at least two thirds.
i3PosNet trained on synthetic images generalizes to real x-rays without any further adaptation. 

\textbf{Conclusion}:
The translation of Deep Learning based methods to surgical applications is difficult, because large representative datasets for training and testing are not available.
This work empirically shows \changed{sub-millimeter pose estimation trained} solely based on synthetic training data.

\keywords{instrument pose estimation \and modular deep learning  \and fluoroscopic tracking  \and minimally invasive bone surgery  \and cochlear implant  \and vestibular schwannoma removal}
\end{abstract}

\section{Introduction}

Current clinical practice in temporal bone surgery for cochlear implantation (CI) and 
vestibular schwannoma removal is still centered on a conventional and open operation setting. 
One fundamental challenge in moving to less traumatic minimally-invasive procedures 
is to satisfy the required navigation accuracy. 
To ensure delicate risk structures such as the facial nerve and chorda tympani are not damaged by surgical tools, the clinical navigation accuracy must exceed \SI{0.5}{\milli\meter} \cite{Schipper.2004,Labadie.2014}.
Recent efforts have used force-feedback \cite{Williamson.2013}, optical tracking systems (OTSs) \cite{Caversaccio.2017} and neuro-monitoring for CI \cite{Anso.2016}, but each one of these strategies have drawbacks in the minimally-invasive setting.
For example OTSs require line-of-sight and a registration between patient and tracking system.
Electromagnetic tracking \cite{Kugler.2019}, force-feedback or neuro-monitoring on the other hand, feature limited accuracy. 
None of these methods can be used to navigate next-generation flexible instruments that follow non-linear paths \cite{Fauser.2018}.
X-ray imaging, on the other hand, is precise and not constrained by line-of-sight.
However, similar to OTS, fiducials used for patient registration significantly impact tracking accuracy of surgical instruments.
The small size of fiducials, 
low contrast to anatomy alongside high anatomy-to-anatomy contrast 
and rotational symmetry characterize challenges specific to pose estimation of surgical tools in temporal bone surgery. 

Unlike previous methods, Deep Learning allows instrument pose estimation to break into submillimeter accuracy at acceptable execution times \cite{Miao.2016,Esfandiari.2018}.
Previous non-Deep Learning pipelines based on 2D/3D registration \cite{Kugler.2018b} and template matching \cite{Vandini.2017b} achieve submillimeter accuracy for simple geometries.
However, such techniques do not scale, do not generalize to more complex instruments and/or require full-head pre-operative CT scans.
In addition, these solutions are usually customized to a specific application, e.g. screws in pedicle screw placement \cite{Esfandiari.2018}, fiducials \cite{Jain.2005b} or guide-wires and instruments \cite{Vandini.2017b}.
The recent shift to Deep Neural Networks offers better accuracy at near real-time execution speed for implants and instruments \cite{Miao.2016,Esfandiari.2018}.
However, no such solution has been proposed for temporal bone surgery.

We propose i3PosNet, a Deep-Learning-powered Iterative Image-guided Instrument Pose Estimation method to provide high-precision estimation of poses.
We focus on pose estimation from x-ray images showing a fiducial (screw) placed on the skull close to the temporal bone, because screws are commonly used as fiducials in robotic CI \cite{Caversaccio.2017}. 
For optimum performance, i3PosNet implements a modular pipeline consisting of 1)~region of interest normalization, 
2)~2D pose estimation and 
3)~3D pose reconstruction to determine the pose.
To this effect, we patchify and normalize a small region around the expected fiducial position.
We design a convolutional neural network (CNN) to predict six pseudo-landmarks on \changed{two axes} with subpixel accuracy.
A geometric model reconstructs 3D poses from their landmark coordinates.
The division of the pipeline into three modular steps reduces complexity, increases performance and significantly boosts the angle estimation performance of i3PosNet.

As part of this work we publish three Datasets in addition to the source code\footnote{Find additional information at \url{https://i3posnet.david-kuegler.de/}}. 
Based on these three Datasets, we show i3PosNet i)~generalizes to real x-ray while only training on synthetic images, 
ii)~generalizes to two surgical instruments in pilot evaluations, and 
iii)~outperforms state-of-the-art registration methods as well as the end-to-end variation of i3PosNet.
As no public datasets with ground truth poses 
are available for training and evaluation of novel pose estimation methods, we believe these new Datasets will foster further developments in this area.
Dataset A consists of synthetic radiographs with a medical screw for training and evaluation.
In Dataset B the screw is replaced by either a drill or a robot.
Dataset C features real images of micro-screws placed on a head phantom. 
Images are acquired with a c-arm and manually annotated in 2D.

\begin{figure}[t]
	\subfloat[\label{fig:projection}three instruments: screw, drill and robot]{\includegraphics[width=60mm]{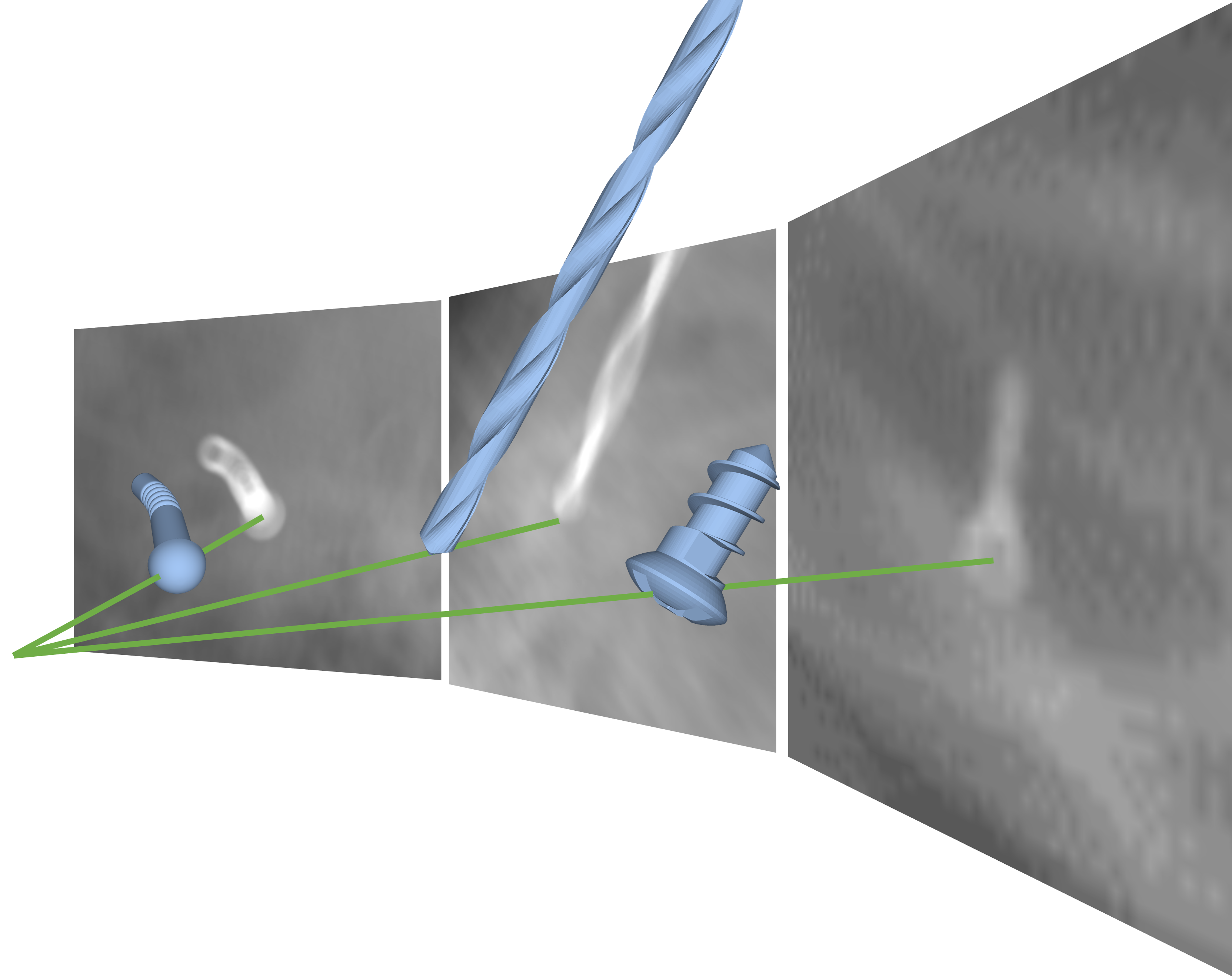}}
	\subfloat[\label{fig:posedefinition}Definition of pose; length not to scale]{\includegraphics[width=60mm]{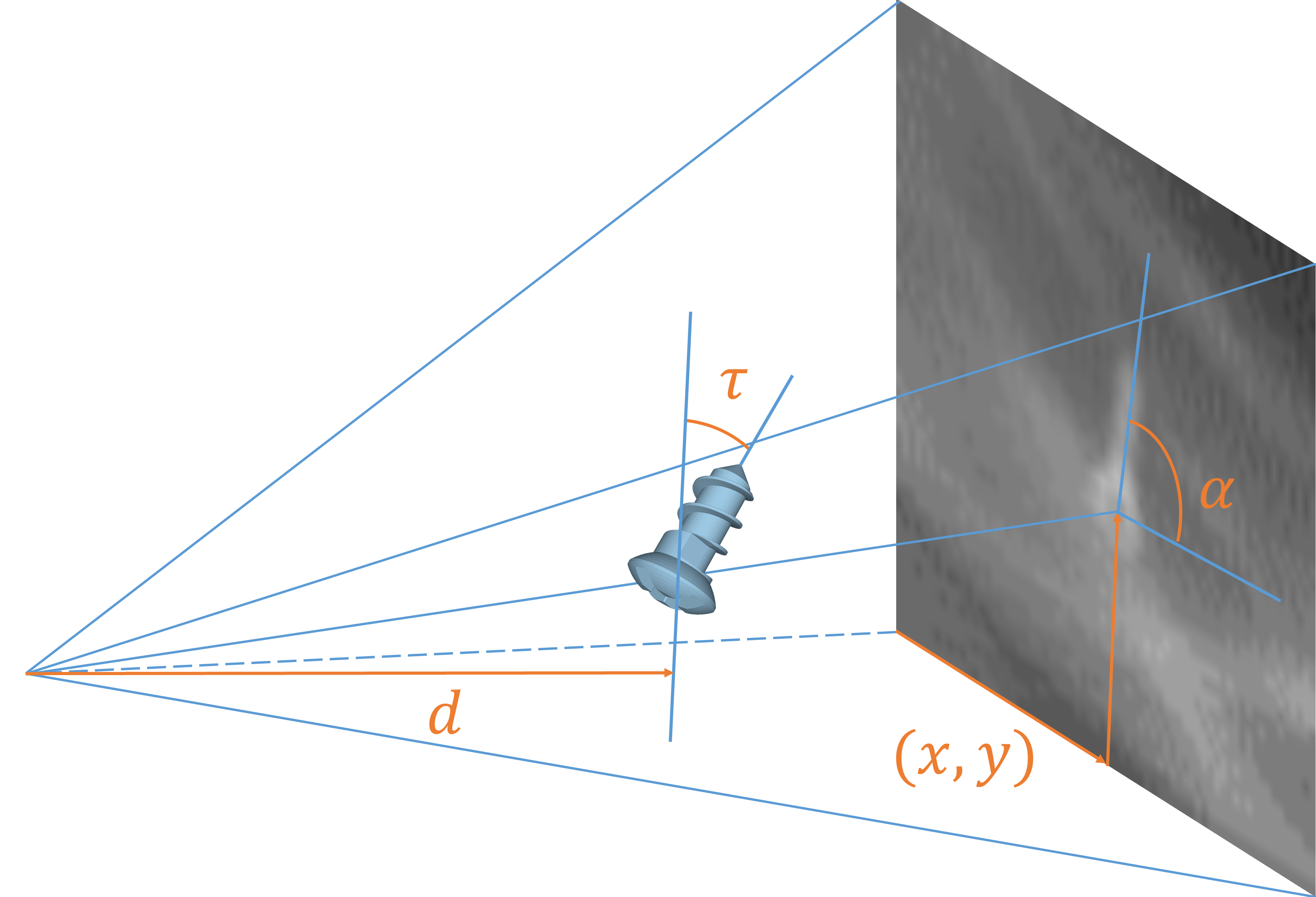}}
	\caption{Instrument Pose Estimation from single x-ray}
\end{figure}

\section{Related Work}
\label{sec:relatedwork}
Pose estimation using radiographs of instruments in the temporal bone has received little scientific attention.
Most published research uses other tracking paradigms, most notably optical tracking \cite{Caversaccio.2017}.
While some Deep-Learning based approaches directly \cite{Bui.2017, Miao.2016} or indirectly \cite{Esfandiari.2018} extract instrument poses from x-ray images, neither address temporal bone surgery.
In this Section, we give a brief overview of instrument tracking for temporal bone surgery.

Robotic solutions for minimally invasive cochlear implantation demand accurate tracking solutions.
The small size of critical risk structures limits the acceptable navigation accuracy to which the tracking accuracy is a significant contributor \cite{Schipper.2004}.
In response, robotic cochlear implantation solutions rely on high precision optical \cite{Caversaccio.2017,Labadie.2014,Balachandran.2010} and electromagentic tracking \cite{Nguyen.2011}.
However, electromagnetic tracking is not suitable to guide a robot because of accuracy and metal-distortion constraints \cite{Kugler.2019}.
Optical tracking cannot directly measure the instrument pose due to the occluded line-of-sight.
Instead, the tool base is tracked adding the base-to-tip registration as a significant error source \cite{Caversaccio.2017}.
\changed{Adding additional, redundant approaches based on
pose reconstruction from bone density and drill forces \cite{Williamson.2013} as well as a neuro-monitoring-based fail-safe 
\cite{Anso.2016},
Rathgeb \etal \cite{Rathgeb.2018} report a navigation accuracy of \SI{0.22 \pm 0.1}{\milli\meter} at the critical facial nerve / \SI{0.11 \pm 0.08}{\milli\meter} at the target.}

Earlier work on x-ray-imaging-based instrument pose estimation centers on traditional registration \cite{Kugler.2018b,Hatt.2016,Jain.2005b,Uneri.2015,Zhang.2017,Gao.2010}, segmentation \cite{Jain.2005b} and template-matching \cite{Vandini.2017b}.
These methods are employed for various applications from pedicle screw placement to implant and guide-wire localization.
Temporal bone surgery has only been addressed by Kügler \etal \cite{Kugler.2018b}.
Recent work introduces Deep Learning methods for instrument pose estimation using segmentation as intermediate representations \cite{Esfandiari.2018} and directly estimating the 3D pose \cite{Miao.2016,Bui.2017}.
While Miao \etal \cite{Miao.2016} employ 974 specialized neural networks, Bui \etal's work \cite{Bui.2017} extends the PoseNet architecture, but does not feature anatomy.
Bier \etal \cite{Bier.2018} proposed an anatomical landmark localization method.
However instruments sizes are significantly smaller in temporal bone surgery impacting x-ray attenuation and therefore image contrast.
For the pose estimation of surgical instruments on endoscopic images \cite{Kurmann.2017,Hajj.2018} Deep Learning is prevalent technique, but sub-pixel accuracy is not achieved -- in part because the manual ground truth annotation do not allow it.

No Deep Learning-based pose estimation method addresses temporal bone surgery or its challenges such as very small instruments and low contrast.
\section{Datasets}
\label{sec:datasets}
This paper introduces three new Datasets: 
two synthetic Digitally Rendered Radiograph (DRR) Datasets (Dataset A for a screw and Dataset B for two surgical instruments), and 
a real x-ray Dataset (Dataset C for manually labeled screws).
All datasets include annotations for the pose with a unified file format.

\subsection{Dataset A: Synthetic}
\label{sec:synthetic}
This dataset shows a CT scan of a head and a screw rendered as x-ray. 
We balance it w.r.t. anatomical variation and projection geometry by implementing a statistical sampling method, which we describe here.
In consequence, the dataset is ideal for method development and the exploration of design choices. 

\textbf{Anatomy and Fiducial}:
To account for the variation of patient-specific anatomy, we consider three different conserved human cadaver heads captured by a SIEMENS SOMATOM Definition AS+.
The slices of the transverse plane are centered around the otobasis and include the full cross-section of the skull.
A small medical screw is virtually implanted near the temporal bone similar to the use for tool registration of the pre-operative CT in robotic cochlear surgery \cite{Caversaccio.2017}.
The screw geometry is defined by a CAD mesh provided by the screw manufacturer (c.f. Dataset C).
Its bounding box diagonal is \SI{6.5}{\milli\meter}.

\begin{table}[t]
	\centering
	\begin{tabular}{l|c|c|c|c|c}
		& object & head anatomy & image generation & annotation & dataset size \\ \hline
		Dataset A & screw & 3 CT scans & synthetic (DRR) & geometric  & 18k images\\ 
		\rowcolor{lightgray!20}Dataset B & drill & 3 CT scans & synthetic (DRR) & geometric  & \by{2}{~18}k\\
		\rowcolor{lightgray!20}& robot & & & & images \\ 
		Dataset C & screw & phantom & real (c-arm) & manual & 540 images\\
		
	\end{tabular}
	\caption{\label{tab:datasets}Dataset Summary}
\end{table}

\textbf{Method for Radiograph Generation}:
\label{sec:Generation}
Our DRR Generation pipeline is fully parameterizable and tailored to the 
surgical pose estimation use-case.
We use the Insight Segmentation and Reconstruction Toolkit 
and 
the Registration Toolkit 
to modify and project the CT anatomy and the fiducial into 2D images.
The pipeline generates projections and corresponding 
ground truth poses from a CT anatomy, a mesh of the fiducial
and a parameter definition, where most parameters can be defined statistically. 
Since the CT data only includes limited sagittal height, we require all projection rays that pass through an approx. \SI{5}{\milli\meter} sphere around the fiducial to be uncorrupted by regions of missing CT data.
For this, missing data is delineated by polygons.
We export the pose $\formulapose$ (\eqnref{eqn:position}) of the instrument for later use in training and evaluation.

\textbf{Parameters of Generation}:
The generation process is dependent on the projection geometry and the relative position and orientation of the fiducial w.r.t. the anatomy.
In specific, we manually choose 10 realistic fiducial poses w.r.t. the anatomy per \changed{subject} and side.
To increase variety, we add noise to these poses (position $\formulainstrumentposition$ and orientation $\formulainstrumentorientation$): 
The noise magnitude for testing is lower than for training to preserve realism and increase generalization.
The projection geometry on the other hand describes the configuration of the mesh w.r.t. x-ray source and detector.
These parameters are encoded in the  Source-Object-Distance, the displacement orthogonal to the projection direction and the rotations around the object.
We derive these from the specification of the c-arm, which we also use for the acquisition of real x-ray images.

In total, the dataset contains 18,000 images across three anatomies. Three thousand of them are generated with the less noise setting for testing.
\subsection{Dataset B: Surgical Tools}
\label{sec:datasteinstruments}
To show generalization to other instruments, we introduce a second synthetic dataset.
This dataset differs from the first two-fold: instead of a screw we use a medical drill or a prototype robot (see \figref{fig:projection}); these are placed at realistic positions inside of the temporal bone instead of on the bone surface.
This dataset includes the same number of images per instrument as Dataset A.

\textbf{Drill}:
Despite the drill's length (drill diameter \SI{3}{\milli\meter}), for the estimation of the tip's pose only the tip is considered to limit the influence of drill bending. 

\textbf{Prototype Robot}:
The non-rigid drilling robot consists of a spherical drilling head and two cylinders connected by a flexible joint. 
By flexing and expanding the joint in coordination with cushions on the cylinders, the drill-head creates non-linear access paths. 
With a bounding box diagonal of the robot up to the joint of \SI{13.15}{\milli\meter}, its dimensions are in line with typical MIS and temporal bone surgery applications. 


\begin{figure}[t]
	\centering%
\includegraphics[width=40mm]{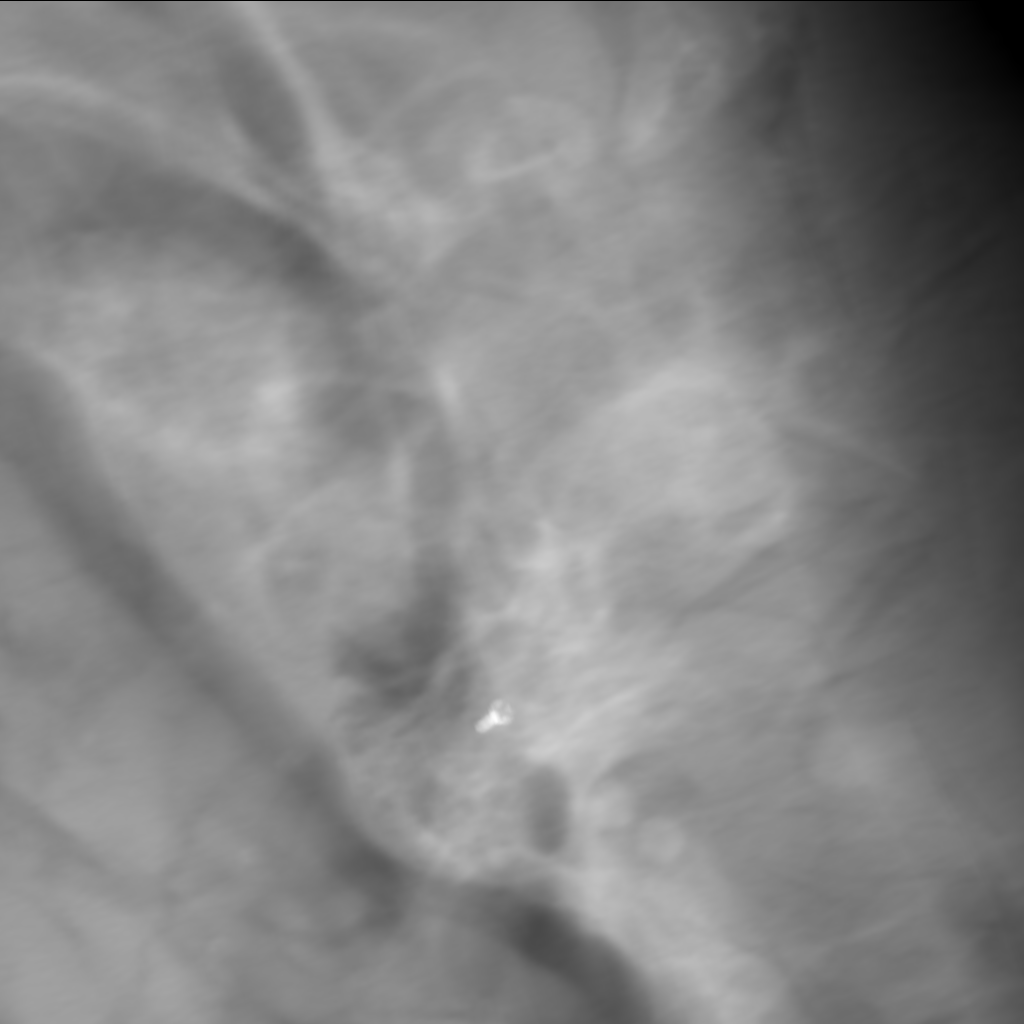}%
\includegraphics[width=40mm]{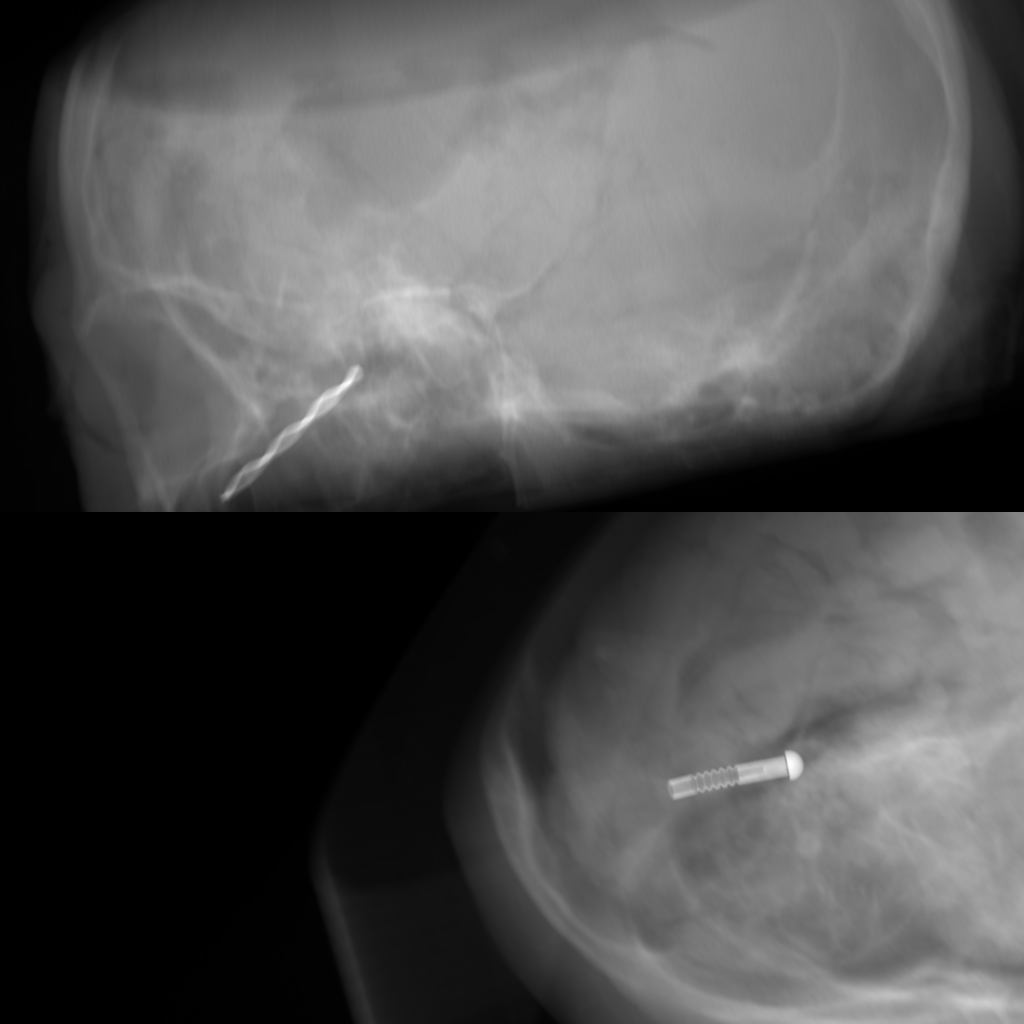}%
\includegraphics[width=40mm]{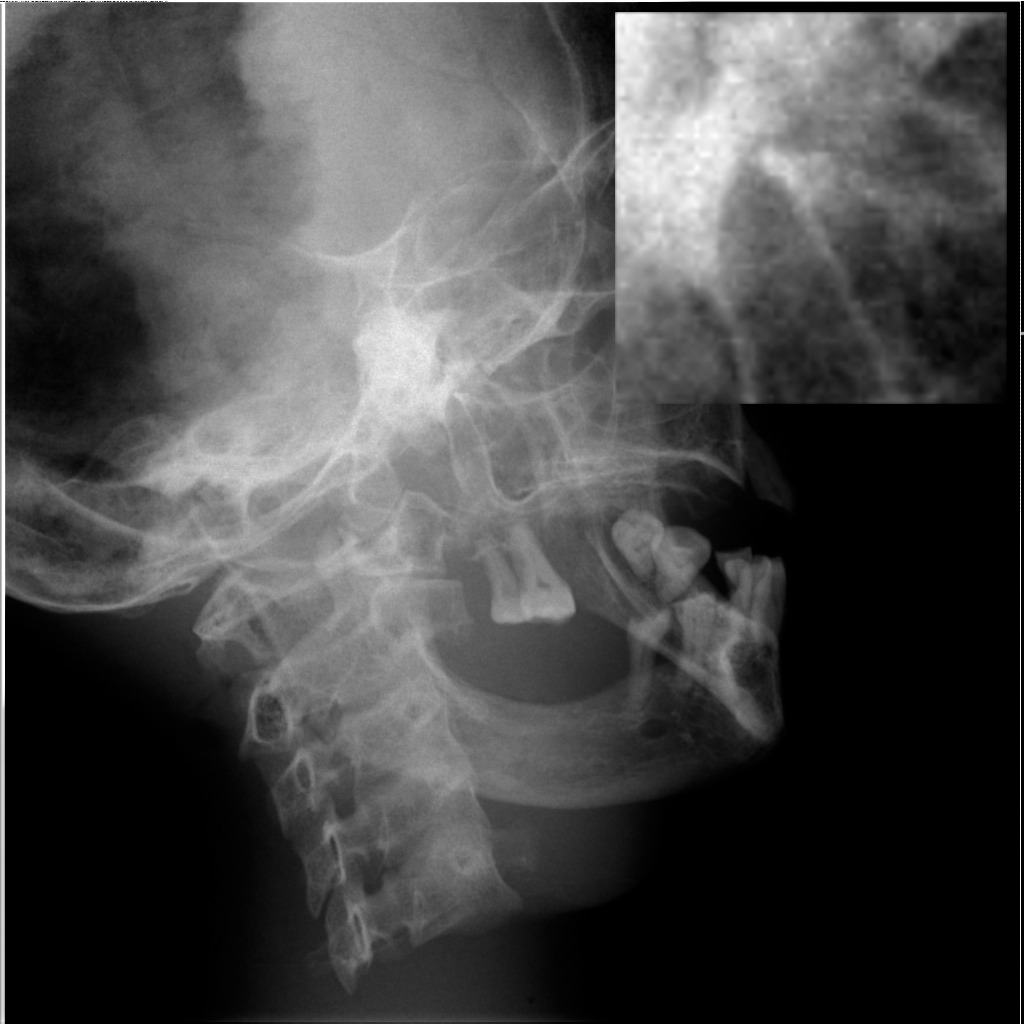}%
\caption{\label{fig:datasets}\changed{Sample images from Dataset A (left), Dataset B (center) and Dataset C (right, the normalized detail illustrates low contrast)}}

\end{figure}

\subsection{Dataset C: Real x-rays}
\label{sec:realxrays}
For Dataset C we acquire real x-ray images and annotations.
The dataset includes 540 images with diverse acquisition angles, distances and positions.

\textbf{Preparation}: 
The experimental setup is based on a realistic x-ray head phantom featuring a human skull embedded in tissue equivalent material.
To not damage the head phantom for the placement of the fiducial (screw), we attach the medical titanium micro-screw with modeling clay as the best non-destructive alternative.
The setup is then placed on the carbon fiber table and supported by x-ray-translucent foam blocks.

\textbf{Image Acquisition}:
We capture multiple x-ray images with a Ziehm c-arm certified for clinical use before repositioning the screw on the skull.
These images are collected from multiple directions per placement.

\textbf{Manual Annotation}:
In a custom annotation tool, we recreate the projection geometry from c-arm specifications replacing x-ray source and detector with camera and x-ray respectively.
The screw is rendered as an outline and interactively translated and rotated to match the x-ray.
We remove images, where the screw is not visible to the annotater (approx. 1\% of images).
Finally, the projected screw position is calculated to fit the projection geometry.
\section{Methods}
\label{sec:methods}
Our approach breaks the task of surgical pose estimation down into three steps: reduction of image variety (region of interest appearance normalization), a convolutional neural network for information extraction and finally pose reconstruction from pseudo-landmarks.
We describe a formalized problem definition in addition to these individual three steps in this section.

\subsection{Problem Definition}
\label{sec:problemdefinition}
Surgical Pose estimation is the task of extracting the pose from a x-ray image.
We define the \emph{pose} $\formulapose$ to be the set of coordinates, \orientationangle{}, projection angle and distance to the x-ray source (depth). 
It is defined w.r.t. the detector and the projection geometry (see \figref{fig:posedefinition}).
\begin{equation}
\formulapose = \left (  x , y , \formulaorientangle, \formulatilt , \formuladepth  \right )^T
\label{eqn:position}
\end{equation}
The \emph{\orientationangle} $\formulaorientangle$ indicates the angle between the instrument's main rotational axis projected onto the image plane and the horizontal axis of the image.
The \emph{projection angle} $\formulatilt$ quantifies the tilt of the instrument w.r.t. the detector plane.
The \emph{depth} $\formuladepth$ represents the distance on the projection normal from the source (focal point) to the instrument (c.f. Source-Object-Distance).

We assume an initial pose with accuracy $\Delta x_\text{initial} \le \SI{2.5}{\milli\meter}$ and $\Delta \alpha_\text{initial} \le \ang{30}$ is available. 
The initial pose can be manually identified, adopted from an independent low-precision tracking system, previous time points or determined from the image by a localization U-Net \cite{Kurmann.2017}.

Given a new predicted pose, the initial guess is updated and the steps described in Sections \ref{sec:appearancenormalization} to \ref{sec:posereconstruction} iteratively repeated. 

\subsection{Appearance Normalization}
\label{sec:appearancenormalization}
Since megapixel x-ray images (\by{1024}{1024} pixels) are dominated by regions unimportant to the pose estimation, 
the direct extraction of subpixel accuracy from megapixel images oversaturates the deep learning strategy with data. 
It is therefore crucial to reduce the size of the image and the image variety based on prior knowledge.

To reduce image variety, the appearance normalization creates an image patch that is rotated, translated and cut to the estimated object position.
Additionally, we normalize the intensity in the patch of \by{92}{48} pixels.
Based on this normalization step and given perfect knowledge of the pose, the object will always appear similar w.r.t. the in-plane pose components (position and \orientationangle).
We define this as \emph{standard pose} -- the object positioned at a central location and oriented in the direction of the x-axis of the patch.

\begin{figure}[t]
	\centering
	\parbox{68mm}{%
		\includegraphics[width=70mm]{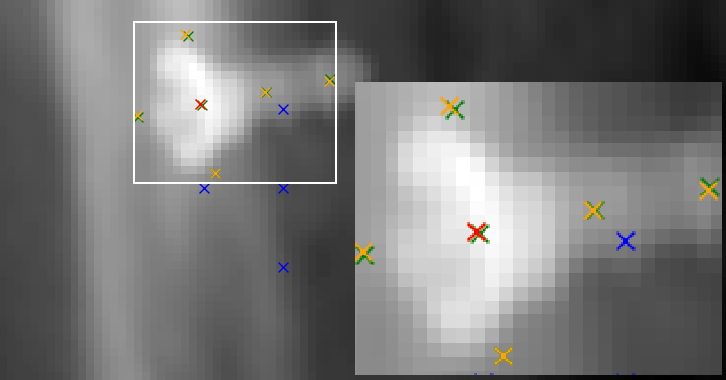}
	}
	~
	\begin{minipage}{47mm}%
		\centering%
		\includegraphics[height=48mm]{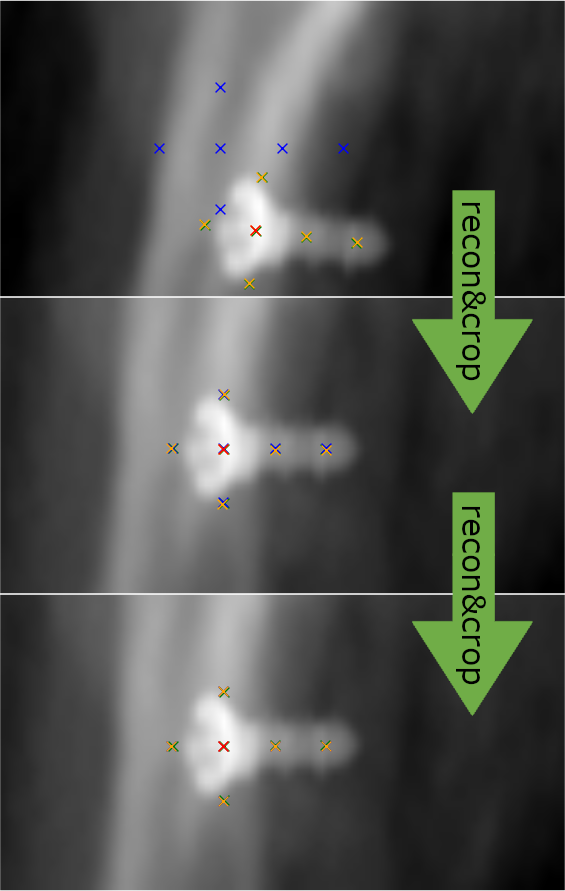}	
	\end{minipage}
	
	\parbox{68mm}{%
		\caption{\label{fig:landmarks}\changed{Pseudo-landmark placement: initial (blue), estimation (yellow) and central landmark (red), ground truth (green)}}}
	~
	\begin{minipage}{47mm}%
		\caption{\label{fig:iteration}\changed{Iterative refinement scheme (recon\&crop: reconstruct pose and crop according to estimation)}}
	\end{minipage}
\end{figure}

Pseudo-landmarks are generated from the pose annotation.
Their geometric placement w.r.t. the fiducial is described by the pair of x- and y-coordinates $(x_i, y_i)$.
The independence from the fiducial's appearance motivates the term ``pseudo-landmark''.
\changed{They are placed \SI{15}{\milli\meter} apart in a cross-shape centered on the instrument (see \figref{fig:landmarks}).
Two normalized support vectors define the legs of the cross: instruments rotational axis ($x$-direction, 3+1 landmarks), and its cross product with the projection direction ($y$-direction, 2+1 landmarks).}
\eqnref{eqn:keypoint2image} formalizes the transformation of the pose to point coordinates in the image plane dependent on the projection geometry ($c_{\text{d2p}} = {\Delta_\text{ds} / d_\text{SDD}}$ with  Source-Detector-Distance $d_{SDD}$ and Detector-Pixel-Spacing $\Delta_{ds}$).
\changed{$\left \{(x^\text{LP}_i , y^\text{LP}_i)\right \}_{i=1}^6$ describe the local placement (LP) of pseudo-landmarks w.r.t. the support vectors.
Finally, landmark positions are normalized w.r.t. maximum values.}
\begin{equation}\label{eqn:keypoint2image}
(x_{i},y_{i})^T = (x,y)^T +  {1 \over c_{\text{d2p}} \formuladepth} \cdot  R(\formulaorientangle) (x^\text{LP}_i \cdot cos(\formulatilt), y^\text{LP}_i)^T
\end{equation}

To construct prior knowledge for training, we generate random variations of the in-plane components of the pose effecting both image and annotation:
\begin{equation}\label{eqn:deltadraw}\arraycolsep=1pt
\begin{array}{rl}
(R, \beta) \sim &(\uniform{0}{\Delta x_\text{initial}},  \uniform{\ang{0}}{\ang{360}})\\ 
\Delta\formulaorientangle \sim &\gauss[0]{(\frac{1}{3} \Delta \formulaorientangle_\text{initial})^2}
\end{array}
\end{equation}
By drawing the variation of the position $\formulainstrumentposition$ in polar coordinates $(R,  \beta)$ \changed{($\mathcal{U}$: uniform distribution)} and the \orientationangle{} $\Delta \formulaorientangle$ from a normal distribution \changed{($\mathcal{N}$)}, 
we skew the training samples towards configurations similar to the standard pose.
This skew favors accuracy based on good estimates over more distant cases, similar to class imbalance.
In effect, this appearance normalization increases the effectiveness (performance per network complexity) of the Deep Neural Network through the utilization of data similarity.

The patch appearance is dominated by the difference between the actual and the standard pose, i.e. the error of the prior knowledge.
\subsection{Pseudo-Landmark Prediction}
Based on the \by{92}{48}-normalized greyscale patch, a convolutional neural network (CNN) extracts the pose information.
Our analysis shows pseudo-landmarks outperform direct pose prediction.
While we explain the design of the pseudo-landmark estimation here, the direct pose estimation follows a similar strategy.

The CNN is designed after a VGG-fashion \cite{Simonyan.2015} with 13 weight layers. 
We benchmark the CNN on multiple design 
dimensions including the number of convolutional layers and blocks, the pooling layer type, 
the number of fully connected layers and the regularization strategy. 
In this context, we assume a block consists of multiple convolutional layers and ends in a 
pooling layer shrinking the layer size by a factor of \by{2}{2}.
All layers use ReLU activation. 
We double the number of channels after every block, starting with 32 for the first block. 
We use the Mean Squared Error as loss function.
For optimizers, we evaluated both Stochastic Gradient Descent with Nesterov Momentum 
update and Adam including different parameter combinations.

\subsection{Pose Reconstruction}
\label{sec:posereconstruction}
We reconstruct the pose from the pseudo-landmarks based on their placement in a cross-shape ($x^\text{LP}_i = 0$ or $y^\text{LP}_i = 0$).
It enables us to invert \eqnref{eqn:keypoint2image} geometrically by fitting lines through two subsets of landmarks.
The intersection yields the position $\formulaposition = (x,y)^T$ and the slope the \orientationangle{} $\formulaorientangle$.
The depth $\formuladepth$ and projection angle $\formulatilt$ are determined by using \eqnref{eqn:depthreconstruction} and \ref{eqn:tiltreconstruction} on the same landmark subsets. 

\begin{equation}\label{eqn:depthreconstruction}
\formuladepth = {c_{d2p}}^{-1} \cdot {|y^\text{LP}_i - y^\text{LP}_j| \over |(x_i,y_i)^T - (x_j,y_j)^T|_2}, i \ne j , x^\text{LP}_{i/j} = 0
\end{equation}
\begin{equation}\label{eqn:tiltreconstruction}
cos (\formulatilt) = c_{d2p} \formuladepth \cdot {|(x_i,y_i)^T - (x_j,y_j)^T|_2 \over |x^\text{LP}_i - x^\text{LP}_j|}, i \ne j, y^\text{LP}_{i/j} = 0
\end{equation}

\section{Experiments \& Results}
\label{sec:results}
\label{sec:training}

We performed a large number of different experiments with independent analysis of the position $\formulaposition$, \orientationangle{} $\formulaorientangle$, projection angle $\formulatilt$ and depth $\formuladepth$.
In this Section, we present the common experimental setup, evaluation metrics and multiple evaluations of i3PosNet.
We group in-depth evaluations in three blocks: \ref{sec:comparison-state-of-art} general, \ref{sec:ablationstudy} modular design evaluation and \ref{sec:limitations} limitations.
To streamline the description of the training and evaluation pipeline, we only present differences to the initially described common experimental setup.

\subsection{Common Experimental Setup}
Generalization to unseen anatomies is a basic requirement to any CAI methodology, therefore we always evaluate on an unseen anatomy.
Following this leave-one-anatomy-out evaluation strategy, individual training runs only include 10k images, since 5k images are available for training per anatomy.

\textbf{Training}: 
Based on the training dataset (from Dataset A), 
we create 20 variations of the prior pose knowledge (see \secref{sec:appearancenormalization}).
In consequence, training uses 20 different image patches per \by{1024}{1024} image for all experiments (200k patches for training).
We train the convolutional neural network to minimize the Mean-Squared-Error of the Pseudo-Landmark regression with the Adam optimizer and standard parameters.
Instead of decay, we use a learning rate schedule that uses fixed learning rates for different training stages of $5 \cdot 10^{-3}$ initially and decrease the exponent by one every 35 epochs.
We stop the training after 80 epochs and choose the best-performing checkpoint by monitoring the loss on the validation set.

\textbf{Testing}:
The testing dataset (from Dataset A) features the unseen third anatomy, which we combine with 10 randomly drawn initial poses per image.
We use the strategy presented in \secref{sec:appearancenormalization} to draw initial in-plane poses.
For testing, no out-of-plane components are required a priori.
The prediction is iterated three times, where the prior knowledge is updated with pose predictions each iteration.
Images with projection angles $|\formulatilt| > \ang{80}$ are \changed{filtered out}, because the performance significantly degrades biasing results.
This leads to on average 7864 tests per trained model.
This degradation is obvious given the \changed{ambiguity} that arises, if projection direction and fiducial main axis (almost) align.
We provide an in-depth analysis of this limitation in \secref{sec:limitations}.

\textbf{Metrics}:
We evaluated the in-plane and out-of-plane components of the predicted pose independently using 5 error measures. 
During annotation, we experienced,
out-of-plane components are much harder to recover from single images so we expect it much harder to recover for neural networks as well.

\emph{In-plane components}: 
The \emph{Position Error} (also reprojection distance \cite{vandeKraats.2005}) is the Euclidean Distance in a plane through the fiducial and orthogonal to the projection normal.
It is measured in pixel (in the image) or millimeter (in the world coordinate system).
The relationship between Pixel and Millimeter Position Error is image-dependent because of varying distances between source and object.
The \emph{Forward Angle Error} is the angle between estimated and ground truth orientation projected into the image plane in degrees, i.e. the in-plane angle error.

\emph{Out-of-plane components}: 
For the \emph{Projection Angle Error}, we consider the tilt of the fiducial out of the image plane in degrees.
Since the sign of the projection angle is not recoverable for small fiducials  ($cos(\formulatilt) = cos (-\formulatilt)$), we only compare absolute values for this out-of-plane angle error.
The rotation angle is not recoverable at all for rotationally symmetric fiducials specifically at this size.
Finally, the \emph{Depth Error} (millimeter) considers the distance between x-ray-source and fiducial (also known as the target registration error in the projection direction \cite{vandeKraats.2005}).

\begin{figure}[t]
	\centering
	\parbox{53mm}{%
		\includegraphics[width=55mm]{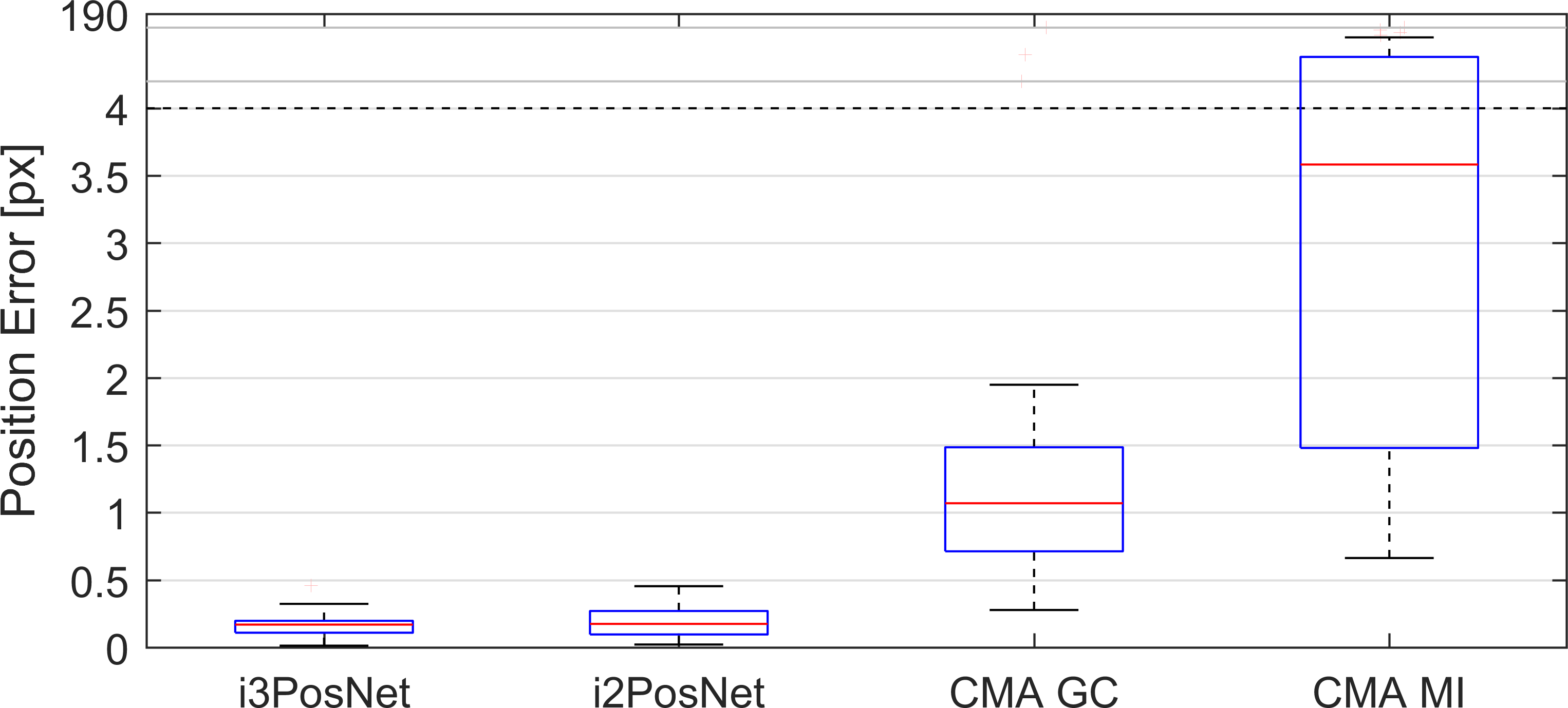}
	}
	~
	\begin{minipage}{60mm}%
		\includegraphics[width=60mm]{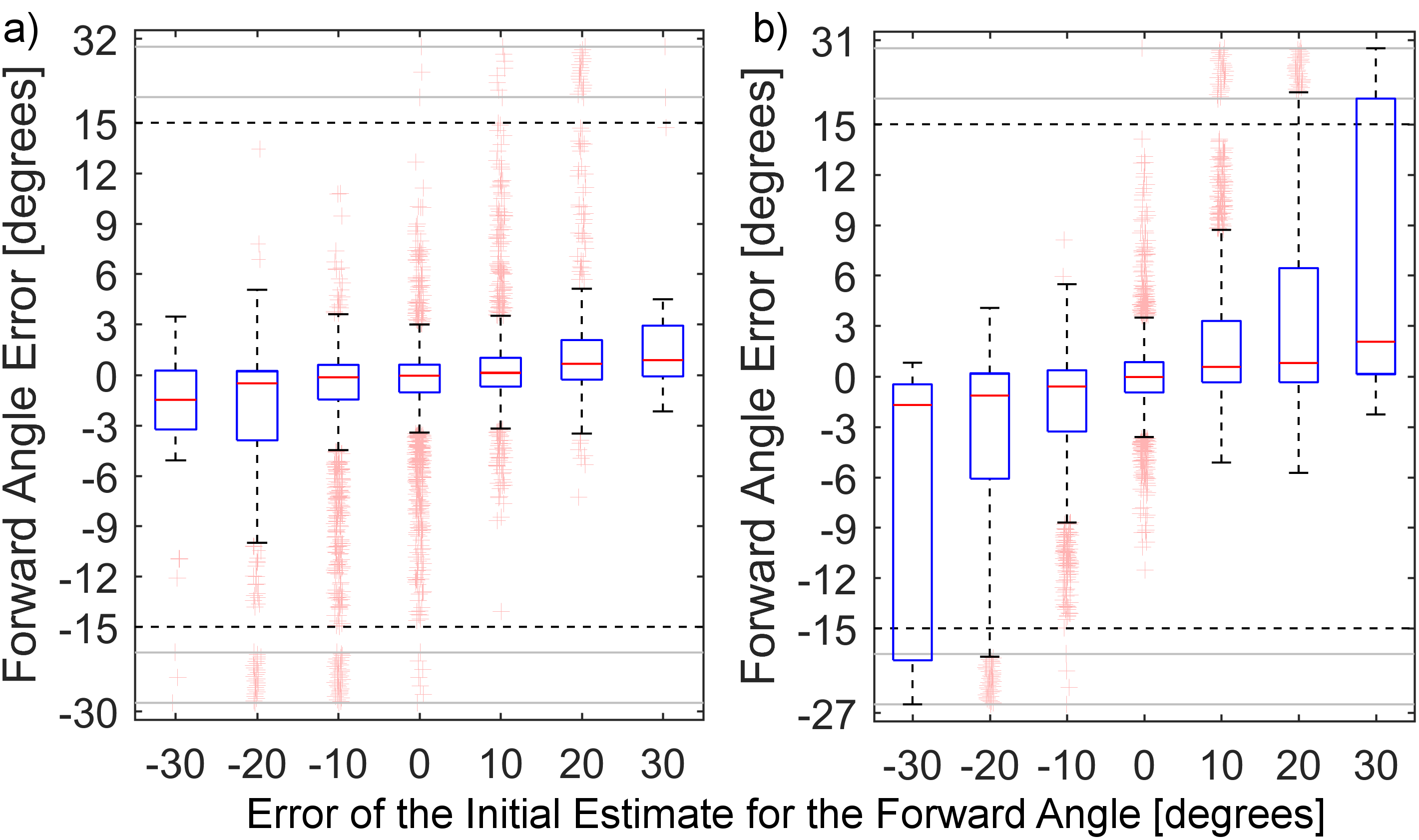}	
	\end{minipage}
	
	\parbox{53mm}{%
		\caption{\label{fig:comparison}Quantative Comparison of i3PosNet, i2PosNet (no iter.), Registration with Covariance Matrix Adaptation Evolution and Gradient Correlation or Mutual Information}}
	~
	\begin{minipage}{60mm}%
		\caption{\label{fig:mvse2e}The addition of virtual landmarks (modular, a) improves \orientationangle{} errors for inaccurate initial angles in comparison to regressing the angle directly (end-to-end, b)}
	\end{minipage}
\end{figure}

\begin{table}[b]
	\centering
	\begin{tabular}{l|r|r|r}
		& Dataset A & Dataset B & Dataset C \\ \hline
		Position Error [mm] & $0.024 \pm 0.016$ & $0.029 \pm 0.028$ & $0.746 \pm 0.818$ \\ 
		\rowcolor{lightgray!20}Position Error [px] & $0.178 \pm 0.118$ & $0.201 \pm 0.178$ & $2.79 \pm 2.89$\\
		Forward Angle Error [deg] & $-0.024 \pm 1.215$ & $0.075 \pm 1.082$ & $6.59 \pm 10.36$ \\
		\rowcolor{lightgray!20}Depth Error [mm] & $1.010 \pm 9.314$ & $0.407 \pm 7.420$ & N/A\\
		Projection Angle Error [deg]& $1.717 \pm 2.183$ & $1.559 \pm 2.130$ & N/A\\
	\end{tabular}
	\caption{\label{tab:results}Results for experiments of synthetic (Dataset A) and real (Dataset C) screw experiments and additional instruments (Dataset B)}
\end{table}

\subsection{Evaluation}
\textbf{Comparison to registration}:
\label{sec:comparison-state-of-art}
Due to long execution times of the registration (\SI{>30}{\minute} per image), the evaluation was performed on a 25-image subset of one (unseen) anatomy with two independent estimates image.
We limited the number of DRRs generated online to 400. 
At that point the registration was always converged.
Both i3PosNet and i2PosNet metrics represent distributions from 4 independent trainings to cover statistical training differences.
Comparing i3PosNet to two previously validated registration methods \cite{Kugler.2018b}, i3PosNet outperforms these by a factor of 5 (see \figref{fig:comparison}).
The errors for i3PosNet and i2PosNet are below $0.5$ Pixel (\SI{0.1}{\milli\meter}) \emph{for all images}.
At the same time i3PosNet reduces the single-image prediction time to \SI{57.6}{\milli\second} on one GTX 1080 at 6\% utilization.

\textbf{Real x-ray image evaluation}:
Because of the significant computation overhead (projection), we randomly choose 25 images from anatomy 1 in Dataset A and performed two pose estimations from randomly sampled deviations from the initial estimate.
Four i3PosNet-models were independently trained for 80 epochs and evaluated for 3 iterations (see \tabref{tab:results}).

\textbf{Generalization}:
i3PosNet also generalizes well to other instruments.
Training and evaluating i3PosNet with corresponding images from Dataset B (drill and robot) shows consistent results across all metrics.

\begin{figure}[t]
	\centering
	\includegraphics[width=100mm]{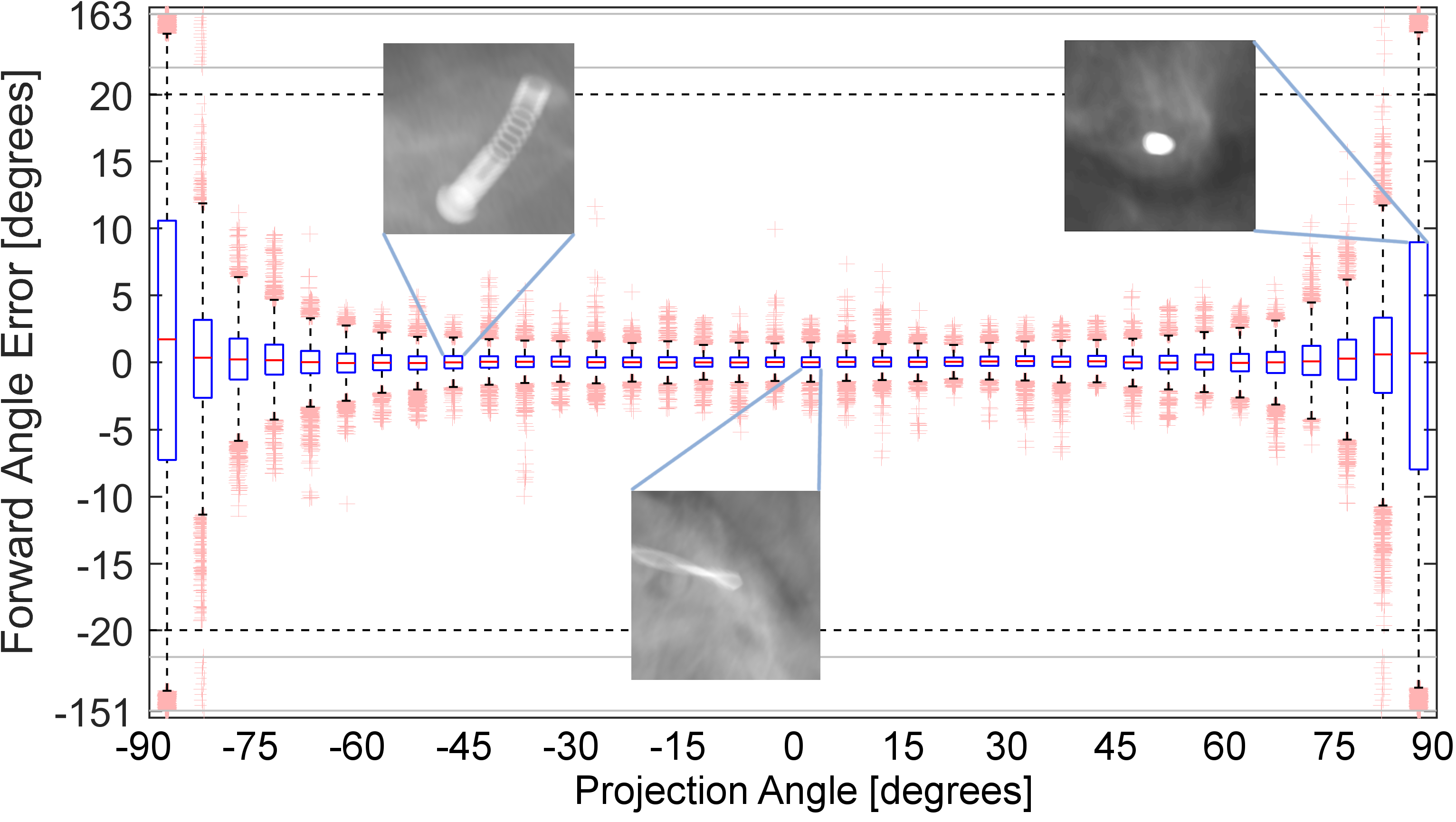}
	\caption{\label{fig:forwardanglebyprojangle}Evaluation of the \orientationangle{} dependent on the projection angle; examples showing different instruments for different projection angles}
	
\end{figure}

\subsection{Evaluation of design decisions}
\label{sec:ablationstudy}
To emphasize our reliance on geometric considerations and our central design decision, 
we evaluated the prediction of forward angles (\emph{end-to-end}) in comparison to the usage of pseudo-landmarks and 2D pose reconstruction (\emph{modular}).
Comparing our modular solution to the end-to-end setup, we found the latter  
to display significantly larger errors for the \orientationangle{}, especially for relevant cases of bad initialization (see \figref{fig:mvse2e}).

\subsection{Limitations to projection parameters}
\label{sec:limitations}
We evaluate the dependence on projection angles $\formulatilt$ imposed especially for drill-like instruments (strong rotational symmetry) (see \figref{fig:forwardanglebyprojangle}).
We observe a decreasing quality starting at \ang{60} with instabilities around \ang{90} motivating the exclusion of images with $|\formulatilt| > \ang{80}$ from the general experiments.

\section{Discussion \& Conclusion}

We estimate the pose of three surgical instruments using a Deep-Learning based approach.
By including geometric considerations into our method, we are able to approximate the non-linear properties of rotation and projection.

\changed{The accuracy provided by i3PosNet improves the ability of surgeons to accurately determine the pose of instruments, even when the line of sight is obstructed.
However, the transfer of the model trained solely based on synthetic data significantly reduces the accuracy, a problem widely observed in learning for CAI \cite{Unberath.2019,Terunuma.2018,Zhang.2018,Vercauteren.2020}.
As a consequence, while promising on synthetic results, i3PosNet closely misses the required tracking accuracy for temporal bone surgery on real x-ray data.
Solving this issue is a significant CAI challenge and requires large annotated datasets mixed into the training \cite{Bui.2017} or novel methods for generation \cite{Unberath.2018}.}

In the future, we also want to embed i3PosNet in a multi-tool localization scheme, where fiducials, instruments etc. are localized and their pose estimated without the knowledge of the projection matrix.
To increase the 3D accuracy, multiple orthogonal x-rays and a proposal scheme for the projection direction may be used.
Through this novel navigation method, surgeries previously barred from minimally-invasive approaches are opened to new possibilities with an outlook of higher precision and reduced patient surgery trauma.

\section*{Compliance with Ethical Standards}\label{CES}

\textbf{Disclosure of potential conflicts of Interest: }
This research was partially funded by the German Research Foundation under Grant FOR 1585. 

\noindent\textbf{Research involving Human Participants and/or Animals: }This article does not contain any studies with human participants or animals. 

\noindent\textbf{Informed consent: }This articles does not contain patient data.


%

\bibliographystyle{spbasic}      

\bibliography{bibliography}
\end{document}